\documentclass{article}

\usepackage[preprint]{neurips_2022}

\usepackage[utf8]{inputenc} 
\usepackage[T1]{fontenc}    
\usepackage{hyperref}       
\usepackage{url}            
\usepackage{booktabs}       
\usepackage{amsfonts}       
\usepackage{nicefrac}       
\usepackage{microtype}      
\usepackage{xcolor}         
\usepackage{epsfig}
\usepackage{multirow}
\usepackage{amsmath,amsfonts,bm}
\usepackage{bbding}
\usepackage{wrapfig}
\usepackage{tablefootnote}

\newcommand{\name}{MRA}
\definecolor{revcolor}{rgb}{1,0,0}

\title{Masked Autoencoders are Robust Data Augmentors}

\author{%
Haohang Xu\textsuperscript{1} \; Shuangrui Ding\textsuperscript{1} \; Manqi Zhao\textsuperscript{2} \;  Dongsheng Jiang\textsuperscript{2} \\
\textsuperscript{1}Shanghai Jiao Tong University \; \textsuperscript{2}Huawei Inc.\\
\small\texttt{\{xuhaohang,dsr1212\}@sjtu.edu.cn}\\
\small\texttt{zhaomanqi97@163.com}\quad\small\texttt{jiangdongsheng1@huawei.com}
}

\begin{document}

\maketitle

\begin{abstract}
Deep neural networks are capable of learning powerful representations to tackle complex vision tasks but expose undesirable properties like the over-fitting issue. To this end, regularization techniques like image augmentation are necessary for deep neural networks to generalize well. Nevertheless, most prevalent image augmentation recipes confine themselves to off-the-shelf linear transformations like scale, flip, and colorjitter. Due to their hand-crafted property, these augmentations are insufficient to generate truly hard augmented examples. In this paper, we propose a novel perspective of augmentation to regularize the training process. Inspired by the recent success of applying masked image modeling to self-supervised learning, we adopt the self-supervised masked autoencoder to generate the distorted view of the input images. We show that utilizing such model-based nonlinear transformation as data augmentation can improve high-level recognition tasks. We term the proposed method as \textbf{M}ask-\textbf{R}econstruct \textbf{A}ugmentation (MRA). The extensive experiments on various image classification benchmarks verify the effectiveness of the proposed augmentation. Specifically, MRA consistently enhances the performance on supervised, semi-supervised as well as few-shot classification.
\end{abstract}

\section{Introduction}
Computer vision has witnessed the mighty power of deep learning over the past decade. 
Through the revolution of backbone models, training datasets, optimization methods. \emph{etc.}, this data-driven learning scheme has achieved major breakthroughs across various vision tasks ranging image classification~\cite{krizhevsky2012imagenet, simonyan2014very, he2016deep}, object detection~\cite{ren2015faster,redmon2016you} and scene segmentation~\cite{long2015fully, chen2017deeplab}. 
However, these methods are heavily dependent on large scale of data to avoid overfitting, where the model perfectly fits the training data via forcibly memorizing the training data~\cite{zhang2021understanding, zhang2018mixup}, but suffers poor performance on the test set. To alleviate the overfitting issue, data augmentations~\cite{lecun1998gradient, krizhevsky2012imagenet} are employed as common training tricks to increase the diversity of training data, especially for small-scale datasets. The mainstream training recipes adopt basic image manipulation as data augmentations~\cite{touvron2021training, liu2022convnet}, which can be mostly represented as linear transformations, consisting of kernel filters, color space transformation, geometric transformation~\cite{shorten2019survey}, \emph{etc.}. These manual-designed methods are fast, reproducible and reliable to encode the invariance of color and geometric space on the original dataset. At the same time, they enjoy the label-preserving property that the transformations conducted over an image would not change the high-level semantic information. However, recent works on self-supervised learning~\cite{gidaris2018unsupervised, zhang2019aet} reveal that these low-level transformations can be easily grasped by the deep neural network, which demonstrates that such basic image processing methods may be insufficient to effectively generalize the input distribution. 

\begin{figure}
    \centering
    \includegraphics[width=0.9\textwidth]{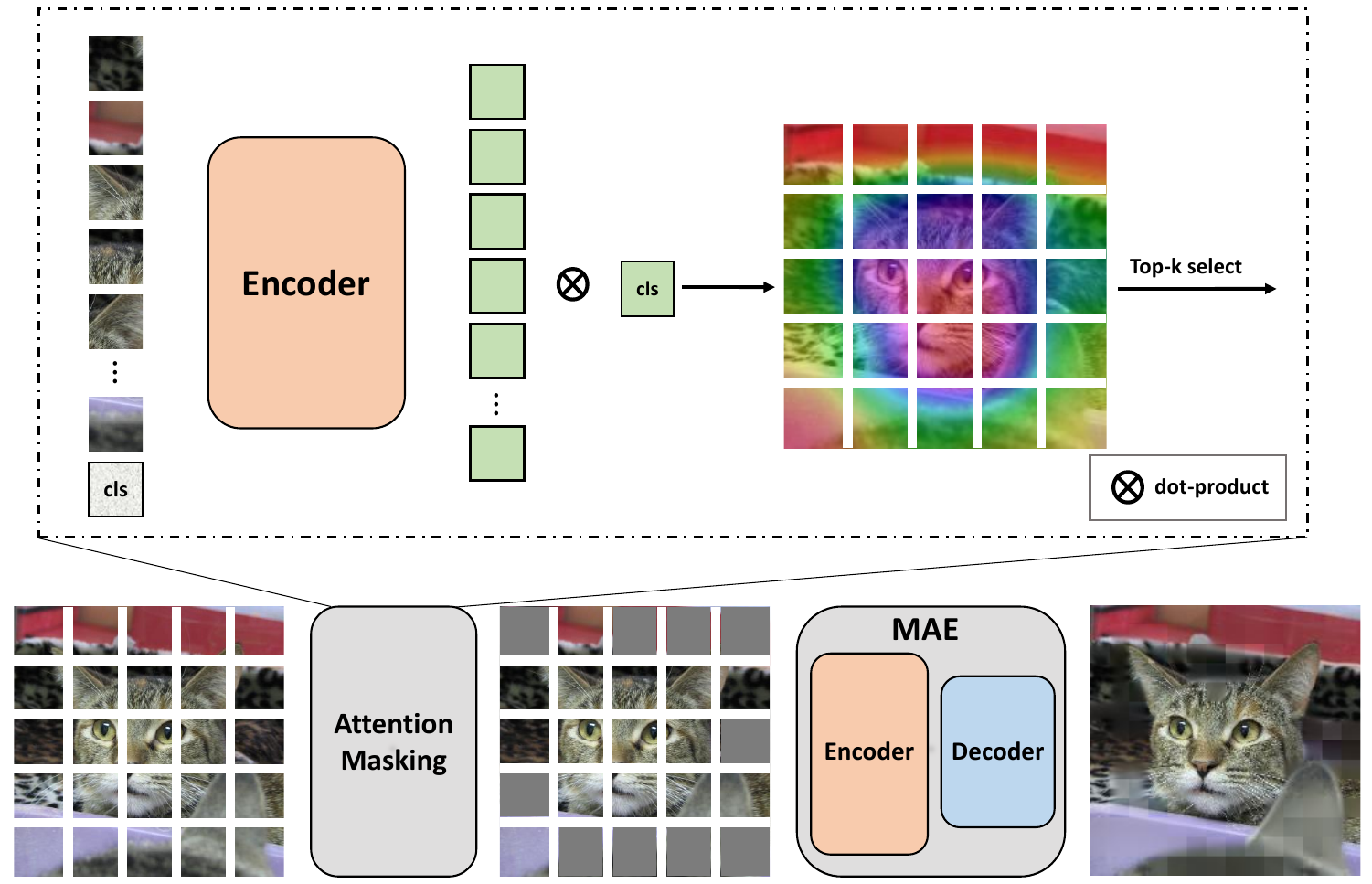}
    \caption{An overview of Mask-Reconstruct Augmentation (MRA). We first mask out the image patches via the attention-based masking strategy. Using class token as query, we calculate attention weights with the keys of each image patch. We remain patches of top-k greatest attention visible. Then, the pretrained masked autoencoder (MAE)~\cite{he2021masked} completes the original image relying on the visible image patches. The reconstructed image can be viewed as robust augmentation for a number of classification tasks, such as supervised, semi-supervised and few-shot classification.}
    \label{fig:overviewarch}

\end{figure}

Instead of using conventional image manipulation,  a line of works introduce generative adversarial networks (GAN)~\cite{goodfellow2014generative, isola2017image} to improve the quality of data augmentations, which can be seen as a type of model-based data augmentation. GAN is powerful to perform unsupervised generation using two adversarial networks, one generates naturalistic images while the other distinguishes fake images from real images. The synthesized image data works well in low-data regime~\cite{antoniou2017data, schwartz2018delta} where collecting datasets is inconvenient, like  the medical imaging~\cite{yi2019generative, frid2018gan, madani2018chest}. But such sample synthesis method can not generalize well to a large-scale labeled dataset~\cite{deng2009imagenet}. The underlying reason might be that there is no guarantee or quantitative evaluation of the generated results~\cite{mikolajczyk2018data}. The composite samples which look good can have dissimilar distribution compared to the original training data. Instead, models that obtain adjacent likelihood can generate unrealistic samples. As a result, the generated objects might be in any ridiculous shape and appearance, differing greatly from what they were previously distributed. 
Therefore, the uncertain and unstable properties of GAN limit its application in image augmentation.
Hence, we need to make the generation more controllable. In this way, we can construct the augmented images reasonably and effectively. 

This paper follows the model-based data augmentation and claims that the generative-based method, can \emph{de facto} improve high-level recognition if constrained in an appropriate way. Motivated by image inpainting, our method, termed as  \textbf{M}ask-\textbf{R}econstruct \textbf{A}ugmentation (MRA), targets at recovering part of the images instead of adversarial learning. 
In detail, we pretrain an extremely light-weight autoencoder via a self-supervised mask-reconstruct strategy~\cite{he2021masked}. Closely following the recent self-supervised method MAE~\cite{he2021masked}, we first divide the images into patches and mask out a set of patches from the input images, meaning only portions of the images are input to the autoencoder. Then, the autoencoder is required to reconstruct the missing patches in the pixel space. Finally, we take the reconstructed image as augmentation for recognition visual tasks. In this way, MRA can not only conduct strong nonlinear augmentation to train robust deep neural networks but also regulate the generation with similar high-level semantics bounded by the reconstruction task. 
To this end, controllable image reconstruction is a good choice for generating a similar likelihood distribution.
In other words, we can generate the robust vicinity of the original image with similar semantics and enable the model to generalize well in different recognition tasks. During the downstream evaluation, we selectively mask out the patches with low attention values, which are more likely to be the background. We show that erasing the label-unrelated noisy patches leads to a more expected and constrained generation, which is highly beneficial to the stable training and enhances the object awareness of the model.
It is worth noting that the whole pretraining process of MRA is label-free and cost-efficient. 
We evaluate MRA on multiple image classification benchmarks.
MRA obtains superior experimental results across the board. 
Specifically, with ResNet-50, solely applying MRA achieves $78.35\%$ ImageNet Top-1 accuracy with $2.04\%$ gain over the baseline. 
Consistent improvements are achieved on fine-grained, long-tail, semi-supervised, and few-shot classification, showing the strong generalization ability of our method.
Moreover, when testing the model to occluded samples, MRA also shows the strong roubustness compared with CutMix~\cite{yun2019cutmix}, Cutout~\cite{devries2017improved} and Mixup~\cite{zhang2018mixup}, manifesting mask autoencoders are robust data augmentors. 

In a nutshell, this paper makes the following contributions:

\begin{itemize}
    \item Inspired by image inpainting, we propose a robust data augmentation method termed MRA to help regulate the training of deep neural networks.
    \item We further constrain the generation by introducing an attention-based masking strategy, which denoises the training and distills object-aware representations.
    \item MRA boosts the performance uniformly among a bunch of classification benchmarks, demonstrating the effectiveness and robustness of MRA. 
\end{itemize}

\section{Related Work}
\paragraph{Model-free Image Augmentation.} The early image augmentations are model-free affine transformations in color and geometric spaces. These transformations encode the invariances of the images which can improve the model performance on image recognition tasks. For example, elastic distortions, translation, scale in MNIST classification~\cite{ciregan2012multi, wan2013regularization}, randomly cropping, flipping, and color jittering in ImageNet and CIFAR classification~\cite{krizhevsky2012imagenet}. 
Moreover, Cutout~\cite{devries2017improved} and Random erasing~\cite{zhong2020random} are designed to overcome occlusion for image recognition challenges. 
Besides, there is another line of work utilizing inter-samples to train the model more robustly.
Mixup~\cite{zhang2018mixup} composes a new image by mixing two different images.
CutMix~\cite{yun2019cutmix} copies a random patch from one image and pastes it into another image, which significantly boosts the robustness and performance. 
Though model-free augmentations are efficient, the difficulty of these augmentations seems to be inadequate for deep model~\cite{gidaris2018unsupervised, zhang2019aet}. 
This paper belongs to the model-based image augmentation, which is discussed next.

\paragraph{Model-based Image Augmentation.} In light of the success of neural architecture search (NAS)~\cite{cai2018proxylessnas}, automated data augmentation methods have made remarkable progress over the past few years. AutoAugment~\cite{cubuk2019autoaugment} proposes to search for optimal combination of each augmentation magnitude. To reduce AutoAugment's high computational cost, FastAA~\cite{lim2019fast} adopts a more efficient policy via density matching. 
FasterAA~\cite{hataya2020faster} and DADA~\cite{li2020dada} perform searching using differential optimization directly, which saves much computational cost. 
RandAugment~\cite{cubuk2020randaugment} reaches the competitive performance by merely setting up two parameters in the same augmentation spaces. 
Besides, thanks to generative adversarial networks~\cite{goodfellow2014generative} and variational auto-encoders~\cite{kingma2013auto}, we can generate new training data that help the model obtain smoother decision boundaries.  
There are works found that augmentations generated by DCGAN~\cite{radford2015unsupervised} and CycleGAN~\cite{isola2017image} can improve the performance of the liver lesion classification~\cite{frid2018gan} and emotion classifcation~\cite{zhu2018emotion}, respectively. 
However, most works applying GANs to image augmentation have been done in biomedical image analysis~\cite{yi2019generative}. 
The direction of generative augmentations remains unexplored on mainstream image recognition benchmarks.
Our work fills the blank, using a masked autoencoder to generate the augmented images. 
\paragraph{Image Inpainting.} Image inpainting~\cite{bertalmio2000image} aims to generate the missing region of an image, which is a crucial problem in computer vision.
To deduce the missing regions, the model needs to grasp the context of an image such as color and texture according to the rest of the image. 
Most inpainting methods follow the pipeline of a context encoder~\cite{pathak2016context}, which infers the missing parts with a generator network using pixel-wise reconstruction loss, and a discriminator to distinguish whether the recovered image is real or fake. 
Image inpainting as a proxy task recently attracts a new wave of self-supervised learning~\cite{he2021masked, bao2021beit, fang2022corrupted, chen2022context, assran2022masked} owing to the success of the vision transformer~\cite{dosovitskiy2020image}.  
In this paper, we closely follow the model architecture of MAE~\cite{he2021masked}. The difference between MAE and MRA is that MAE directly takes the encoder as the ultimate pretrained backbone while MRA leverages the autoencoder to augment the original image and then train another backbone.

\section{Approach}

In this section, we introduce our Mask-Reconstruct Augmentation (\name). In section~\ref{mae},  we first revisit the pretraining framework based on masked autoencoder~\cite{he2021masked}. 
Then, we detail an attention-based masking strategy to constrain the augmentation in section~\ref{att}. Eventually, section~\ref{mra} illustrates our whole pipeline shown in Figure~\ref{fig:overviewarch}. We adopt the pretrained masked autoencoder as the data augmentor to reconstruct masked input images for downstream classification tasks. 


\subsection{Masked Autoencoders}
\label{mae}
Given unlabeled training set ${\bm X}=\left \{\mathbf{x_{1}},\mathbf{x_{2}},...,\mathbf{x_{N}}\right \}$, the masked autoencoder aims to learn an encoder $\mathrm{E}_{\theta}$ with parameters $\theta$ : $\mathbf{M} \odot \mathbf{x} \mapsto \mathrm{E}_{\theta}(\mathbf{M}\odot \mathbf{x})$, where $\mathbf{M} \in \left\{0,1\right\} ^ {W\times H}$ denotes a block-wise binary mask with block size of $16\times 16$ pixels. Meanwhile, we train an decoder $\mathrm{D}_{\phi}$ with parameters $\phi$ to recover original image from masked image's latent embeddings: $\hat{\mathbf{x}} = \mathrm{D}_{\phi} (\mathrm{E}_{\theta}(\mathbf{M}\odot \mathbf{x}))$, where $\hat{\mathbf{x}}$ designates the reconstructed image. We train both encoder $\mathbf{E}_{\theta}$ and decoder $\mathbf{D}_{\phi}$ end-to-end with the learning objective of the mean squared error (MSE) $\ell(\hat{\mathbf{x}}, \mathbf{x})$ between the reconstructed images $\hat{\mathbf{x}}$ and original images $\mathbf{x}$ in the pixel space. In practice, we find significantly squashing the model size of autoencoder remain a considerably high performance, which is reported in Table~\ref{tab:modelsize}. Thus, to strike an ideal balance between the speed and the performance, we devise a mini version of the masked autoencoder, and achieve a throughput of 963 imgs/s on one NVIDIA V100 GPU when integrating it with the ResNet-50 for downstream classification, which is affordable in terms of whole training. We set this mini version of MAE in default for the evaluation. 

\subsection{Attention-based Masking}
\label{att}
To guide the augmentation being object-aware, we leverage the inductive bias of object location into the masking strategy. We adopt attention probing as a reasonable referee to determine whether the patch belongs to the foreground object. We maintain the patches with high attention as input and erase the rest of the patches.
Given the pretrained encoder $\mathrm{E}_{\theta}$, we can compute attention maps for each input patch. 
To fit the input format of vision transformer, the input image $\mathbf{x} \in \mathbb{R}^{H\times W \times C}$ is divided into non-overlapped patches as $\mathbf{x}_p \in \mathbb{R}^{\frac{H}{p}\times\frac{W}{p}\times p^2\times C}$ where ($H$, $W$) represent the height and width of the input image, $C$ denotes the channel dimension, $p$ means the patch size. 
The recent study~\cite{caron2021emerging} has shown that vision transformers trained with no supervision can automatically learn object-related representation. Moreover, the attention map of the class token can provide reliable foreground proposals shown in Figure 1 of~\cite{caron2021emerging}. Driven by this observation, we compute the attention map of the class token on image patch $i$:
\begin{align}
{\mathbf{Attn}}_i = \mathbf{q}_{\rm cls} \cdot \mathbf{k}_i, i \in \left\{0,1,...,p^2 -1 \right\}
\end{align}
where $\mathbf{q}_{\rm cls}$ is the query of class token and $\mathbf{k}_i$ formulates the key embedding of patches $i$. Both $\mathbf{q}_{\rm cls}$ and $\mathbf{k}_i$ are fetched from the last block of encoder. 
Then, we sort the attention map $\mathbf{Attn} = \left\{ {\mathbf{Attn}}_i\  | \ i=0,1,...,p^2-1 \right\}$ and get the top-k index set $\mathbf{\Omega}$:
\begin{align}
\mathbf{\Omega} = \mathbf{top}\mathbf{-}\mathbf{rank}({\mathbf{Attn}}, \mathbf{K})
\end{align}
where function $\mathbf{top}\mathbf{-}\mathbf{rank}(\cdot, \mathbf{K})$ returns the indexes of top-k biggest elements.
Having the top-k index set $\mathbf{\Omega}$, we generate a attention-based binary mask $\mathbf{M}^*$ as:
\begin{align}
\mathbf{M}^* \left[\lfloor\frac{\mathbf{\Omega}_i}{p}\rfloor \frac{H}{p} : \lfloor\frac{\mathbf{\Omega}_i}{p}+1\rfloor\frac{H}{p},\ \  {\rm mod}(\mathbf{\Omega}_i,p)\frac{W}{p} : ({\rm mod}(\mathbf{\Omega}_i,p)+1)\frac{W}{p} \right] = \mathbf{1} \label{eq:mask}
\end{align}
where $\lfloor\cdot\rfloor$ indicates the round down operation, and ${\rm mod (\cdot)}$ implies the modulo operation. 
After applying the attention-based binary mask $\mathbf{M}^*$ on input image $\mathbf{x}$, we expect that the possible background area is effaced, while the foreground area is intact. Noted that we only leverage attention-based masking policy during the downstream tasks,
while we keep randomly masking the patches at the stage of pretraining the autoencoder.

\subsection{Mask-Reconstruct Augmentation}
\label{mra}
The ultimate architecture of MRA is displayed in Figure~\ref{fig:overviewarch}. With the attention-based binary mask $\mathbf{M}^*$, we first acquire the masked image $\mathbf{M}^* \odot \mathbf{x}$. Then, we divide the masked image $\mathbf{M}^* \odot \mathbf{x}$ into non-overlapped patches and discard the masked patches. The remaining unmasked patches are fed into the pretrained encoder ${\rm E}_{\theta}$ and decoder ${\rm D}_{\phi}$ to generate the reconstructed image
$\hat{\mathbf{x}} = {\rm D}_{\phi} ({\rm E}_{\theta}(\mathbf{M}^* \odot \mathbf{x}))$. The reconstructed image $\hat{\mathbf{x}}$ can be seen as an augmented version of $\mathbf{x}$, which can be used in several classification tasks. Note that once pretrained, \name\ is fixed and does not require further finetuning when testing on different datasets and tasks, it can still generate robust and credible augmentation.

\section{Experiments}
In this section, we evaluate \name\ on several classification tasks, including fully supervised, semi-supervised, and few-show classification. Several ablation studies are conducted to diagnose how each component affects the performance. Finally, we test the robustness of \name\ on occluded samples.

\subsection{Pretraining Details}
We pretrain the autoencoder module on ImageNet~\cite{deng2009imagenet} for 200 epochs following the hyper-parameters of MAE~\cite{he2021masked}. As mentioned in section~\ref{mae}, we set up a lightweight autoencoder in default experiments. We term it as MAE-Mini. MAE-Mini stacks 4 layers of the encoder and 2 layers of the decoder with an embedding size of 480. Compared with 12 layers of the encoder and 6 layers of the decoder in the standard MAE setting, MAE-Mini can be integrated into most networks very efficiently.

\subsection{Fully-supervised Image Classification}
\textbf{ImageNet Classification.}
ImageNet~\cite{deng2009imagenet} is a widely used dataset for image classification, which contains 1.2 million training images and 50000 validation images with 1000 classes. The input images are first processed by the standard augmentations such as RandomResizedCrop and flipping. Then the resized 224$\times$224 images are fed into the pretrained \name\ module to perform the mask-and-reconstruct operation. The reconstructed images can be directly exploited to compute classification loss. The standard ResNet-50 is utilized as the backbone. Following the training recipes in~\cite{yun2019cutmix}, the model is trained for 300 epochs using an SGD optimizer with a momentum of 0.9 and weight decay of 0.00006. The batch size and learning rate are set as 512 and 0.3, respectively. As shown in Table \ref{tab: imagenet}, \name\ achieves $78.35\%$ top-1 accuracy using ResNet-50 as backbone, which outperforms a series of automated augmentations searching methods. We also compare the GPU hours of pretraining and pre-searching on ImageNet, \name\ also has an affordable computation cost compared with AutoAugment and Fast AutoAugment. In addition, once pretrained, \name\ can be applied to several classification tasks without additional fine-tuning. Note that CutMix~\cite{yun2019cutmix} and its variants~\cite{uddin2020saliencymix,kim2020puzzle} can achieve the better results by introducing inter-sample regularization. \name\ can also be combined with CutMix to further improve the performance.
By combining CutMix, \name\ achieve $78.93\%$ top-1 accuracy on ImageNet, which outperforms carefully designed mixed strategy~\cite{uddin2020saliencymix, kim2020puzzle}.

\begin{table}[t]
\small
\centering
\caption{ImageNet classification Top-1/5 accuracy based on ResNet-50 backbone. \Checkmark means the method only relies on the intra-sample to conduct augmentation while \XSolidBrush denotes the method exploits the inter-samples. GPU Hours is measured with an NVIDIA V100 GPU. }
\setlength{\tabcolsep}{3.7mm}
\renewcommand\arraystretch{1.}  
\begin{tabular}{lcccc}
\toprule
Method     & Intra-sample? & GPU Hours & Top-1   & Top-5  \\
\midrule
Baseline   &\Checkmark &0 & 76.31     & 92.95       \\
AutoAugment~\cite{cubuk2019autoaugment}  &\Checkmark       &15000 & 77.63     & 93.80          \\
Fast AutoAugment~\cite{lim2019fast} &\Checkmark  &450 & 77.60     &93.70           \\
RandAugment~\cite{cubuk2020randaugment}    &\Checkmark     &0 & 77.60     & 93.80          \\
Faster AutoAugment~\cite{hataya2020faster} &\Checkmark  &2.3 & 76.50     &  93.20         \\
DADA~\cite{li2020dada}  &\Checkmark    &1.3 & 77.50     & 93.50          \\
Cutout~\cite{devries2017improved}  &\Checkmark   &0 & 77.07     &93.34           \\
\name &\Checkmark   &140 & \textbf{78.35}   & \textbf{93.93} 
\\ \midrule
Mixup~\cite{zhang2018mixup} &\XSolidBrush &0 &77.42 &93.60 \\
CutMix~\cite{yun2019cutmix} &\XSolidBrush &0 &78.60 &94.08 \\
SaliencyMix~\cite{uddin2020saliencymix} &\XSolidBrush &280 &78.74 &94.24 \\
Puzzle Mix~\cite{kim2020puzzle} &\XSolidBrush & 576 &78.76 & \textbf{94.29}\\
MRA + CutMix &\XSolidBrush &140 &\textbf{78.93} &94.24 \\
\bottomrule
\end{tabular}
\label{tab: imagenet}
\end{table}

\textbf{Fine-grained Classification.}
We assess the generalization of \name\ on several fine-grained classification datasets, including CUB-200-2011~\cite{wah2011caltech}, FGVC-Aircraft~\cite{maji2013fine} and StanfordCars~\cite{krause20133d}. For all the experiments, we fine-tune the ResNet-50 from the official pretrained checkpoints provided by PyTorch\footnote{https://download.pytorch.org/models/resnet50-19c8e357.pth} for 90 epochs. The learning rate in the SGD optimizer is set as 0.001 and is decayed by 10 every 30 epochs. We keep the hyper-parameters exactly the same during running the baseline supervised experiments and our \name\ experiments to make sure the comparison is fair. As shown in Table \ref{tab:fine-grained}, \name\ consistently improves the performance on fine-grained classification.

\begin{table}[!htb]
    \small
    \caption{Top-1 accuracy on fine-grained datasets. ResNet-50 is used as backbone.}
\renewcommand\arraystretch{1.1}  
    \centering 
    \setlength{\tabcolsep}{3.7mm}
    \begin{tabular}{lccc}
        \toprule
        Method &CUB-200  &Cars &Aircraft \\
        \midrule
        Supervised &84.62 &85.41  &73.45 \\ 
        \midrule
         \name\ &{\bf 85.52} &{\bf 86.45}  &{\bf 76.21} \\
         \bottomrule
    \end{tabular}
    \label{tab:fine-grained}
\end{table}

\begin{table}[!htb]
\small
    \caption{Top-1 test accuracy on long-tail classification. ResNeXt50 is used as backbone.}
\renewcommand\arraystretch{1.1}  
    \centering 
    \setlength{\tabcolsep}{3.7mm}
    \begin{tabular}{lcccc}
        \toprule
        Sampling &Many  &Medium &Few &All \\
        \midrule
        Instance-Balanced~\cite{kang2019decoupling} &65.9 &37.5  &7.7 &44.4 \\ 
        Instance-Balanced + \name\ &\textbf{67.3} &\textbf{39.9} &\textbf{9.6} &\textbf{46.3} \\
        \midrule
        Class-Balanced~\cite{kang2019decoupling} &61.8 &40.1 &15.5 &45.1 \\
        Class-Balanced + \name &\textbf{65.4} &\textbf{41.1} &\textbf{16.3} &\textbf{47.1} \\
         \bottomrule
    \end{tabular}
    \label{tab:long-tail}
\end{table}

\textbf{Long-tail Classification.}
We further evaluate \name\ on long-tail classification. Two balanced sampling methods in~\cite{kang2019decoupling} are used as the baseline: Instance-Balanced and Class-Balanced. The \name\ is directly applied on the $224\times 224$ images after simple RandomResizedCrop augmentation. All the hyper-parameters including the optimizer and epochs are kept the same as the configuration in~\cite{kang2019decoupling} for a fair comparison. ResNeXt50~\cite{xie2017aggregated} is utilized as the backbone for consistency. The detailed configuration can be found in its official github repository\footnote{https://github.com/facebookresearch/classifier-balancing}. As shown in Table \ref{tab:long-tail}, \name\ improves long-tail classification accuracy in two different settings, which verifies its effectiveness. 

\subsection{Semi-supervised Classification}
Semi-supervised classification focuses on label-hungry settings in deep learning. In semi-supervised learning, only a small set of samples are labeled and the rest samples are unlabeled. FixMatch~\cite{sohn2020fixmatch} is a strong baseline method in semi-supervised classification, which proposes to create two augmented versions of one image. Especially, one is processed with a weak augmentation (RandomResizedCrop) and the other is processed with a strong augmentation (RandAugment~\cite{cubuk2020randaugment}). The model is trained to maximize the consistency between two augmented images. The reconstructed image of \name\ can be also seen as a strong augmented version of the original input. We propose to use the reconstructed image of \name\ as a type of strong augmentation in FixMatch. The STL-10 dataset~\cite{coates2011analysis} with 40 labeled and 250 labeled samples are used as two semi-supervised settings. We select WideResNet-28~\cite{zagoruyko2016wide} as the backbone. As shown in Table \ref{tab:semi}, using \name\ augmentation outperforms standard strong augmentation, i.e., RandAugment, in FixMatch obviously, which verifies the effectiveness of \name\ in a different application\footnote{All results in semi-supervised classification are trained with the code-base in https://github.com/TorchSSL/TorchSSL, which is an all-in-one PyTorch toolkit for semi-supervised learning.}.

\begin{table}[!htp]
\small
    \caption{Top-1 accuracy on STL-10 datasets. WideResNet-28 is used as backbone. 40/250 Labels means that only 40/250 labeled samples are available in semi-supervised setting.}
\renewcommand\arraystretch{1.1}  
    \centering 
    \setlength{\tabcolsep}{3.7mm}
    \begin{tabular}{lccc}
        \toprule
        Method &40 Labels  &250 Labels \\
        \midrule
        PiModel~\cite{rasmus2015semi} &25.69 &44.87 \\
        PseudoLabel~\cite{lee2013pseudo} &25.32 &47.94 \\
        VAT~\cite{miyato2018virtual} &25.26 &43.58 \\
        MeanTeacher~\cite{tarvainen2017mean} &28.28 &43.51 \\
        MixMatch~\cite{berthelot2019mixmatch} &45.07 &65.48 \\
        UDA~\cite{xie2020unsupervised} &62.58 &90.28 \\
        FixMatch~\cite{sohn2020fixmatch} &64.03 &90.19 \\ 
        \midrule
        FixMatch + \name\ &{\bf 67.61} &{\bf 92.09} \\
         \bottomrule
    \end{tabular}
    \label{tab:semi}
\end{table}

\begin{table}[!ht]
\small
\centering
\renewcommand\arraystretch{1.1}
\caption{Top-1 accuracy on miniImageNet with $95\%$ confidence interval in few-shot classification. All experiments are from 5-way classification.}
\label{tab:fewshot}
\begin{tabular}{c|ccc}
\toprule
&                  & ResNet-18      & ResNet-34      \\ \midrule
\multirow{6}{*}{\begin{tabular}[c]{@{}c@{}}\\\   1-shot\end{tabular}} 
& {{MatchingNet~\cite{vinyals2016matching}}} & {52.91$\pm$0.88} & {53.20$\pm$0.78} \\
& {{ProtoNet~\cite{snell2017prototypical}}}    & \textbf{{54.16$\pm$0.82}} & \textbf{{53.90$\pm$0.83}} \\
& {{MAML~\cite{finn2017model}}}        & {49.61$\pm$0.92} & {51.46$\pm$0.90} \\
& {{RelationNet~\cite{sung2018learning}}} & {52.48$\pm$0.86} & {51.74$\pm$0.83} \\ 
\cmidrule{2-4}
& {{Baseline~\cite{chen2018closer}}}    & {51.75$\pm$0.80} & {49.82$\pm$0.73} \\ 
& {{Baseline + CutMix~\cite{yun2019cutmix} }} &{53.10$\pm$0.84} &{45.20 $\pm$0.79} \\
& {{Baseline + \name\ }} &{53.16$\pm$0.80} &{51.95$\pm$0.77} \\ \midrule 
\multirow{6}{*}{\begin{tabular}[c]{@{}c@{}}\\   5-shot\end{tabular}}
& {{MatchingNet~\cite{vinyals2016matching}}} & {68.88$\pm$0.69} & {68.32$\pm$0.66} \\
& {{ProtoNet~\cite{snell2017prototypical}}}    & {73.68$\pm$0.65} & {74.65$\pm$0.64} \\
& {{MAML~\cite{finn2017model}}}        & {65.72$\pm$0.77} & {65.90$\pm$0.79} \\
& {{RelationNet~\cite{sung2018learning}}} & {69.83$\pm$0.68} & {69.61$\pm$0.67} \\ 
\cmidrule{2-4}
& {{Baseline~\cite{chen2018closer}}}    & {74.27$\pm$0.63} & {73.45$\pm$0.65} \\
& {{Baseline + CutMix~\cite{yun2019cutmix} }} &{76.60$\pm$0.66} &{68.95 $\pm$0.79} \\
& {{Baseline + \name\ }} &\textbf{{76.71$\pm$0.60}} &\textbf{{75.07$\pm$0.64}} \\ \bottomrule
                                
\end{tabular}
\end{table}

\subsection{Few-shot Classification} 
In few-shot learning, a large number of labeled training samples are given on some base categories first, then the goal is to predict on novel categories where only a few $K$-shot samples are labeled. The base categories and novel categories are not overlapped. We evaluate few-shot classification~\cite{chen2018closer,snell2017prototypical, sung2018learning} on  miniImageNet dataset. MiniImageNet consists of 80 base classes with 600 labeled samples per class, and 20 novel classes with only $K$~($K=1$ or $K=5$) labeled samples per class. Recent work~\cite{chen2018closer} proposed a simple but effective baseline method for few-shot classification, where a backbone is pretrained in a fully supervised fashion on base categories, and a classifier is retrained on novel categories upon the fixed backbone. Based on this baseline, we apply \name\ in the pretraining stage on base categories, and the following retraining stage on novel categories is unchanged. As shown in Table \ref{tab:fewshot}, the model pretrained with \name\  shows a stronger generalization ability on novel categories compared with the baseline method. We also compare the results of pretraining with CutMix augmentation. Interestingly, when pretraining with CutMix on ResNet-34, the performance drops a lot. According to the analysis in~\cite{chen2018closer}, we conjecture the CutMix augmentation leads to a severe over-fitting on base categories, which induces the failure of transferring to novel classes.

                                

\subsection{Ablation Studies}
In this section, we conduct several ablation studies to dissect the effect of each component. Unless specified, we conduct all ablation studies on the ImageNet dataset for 90 epochs using ResNet50 as backbone, and report the Top-1 accuracy of the validation set.

\begin{figure}
    \centering
    \includegraphics[scale=.42]{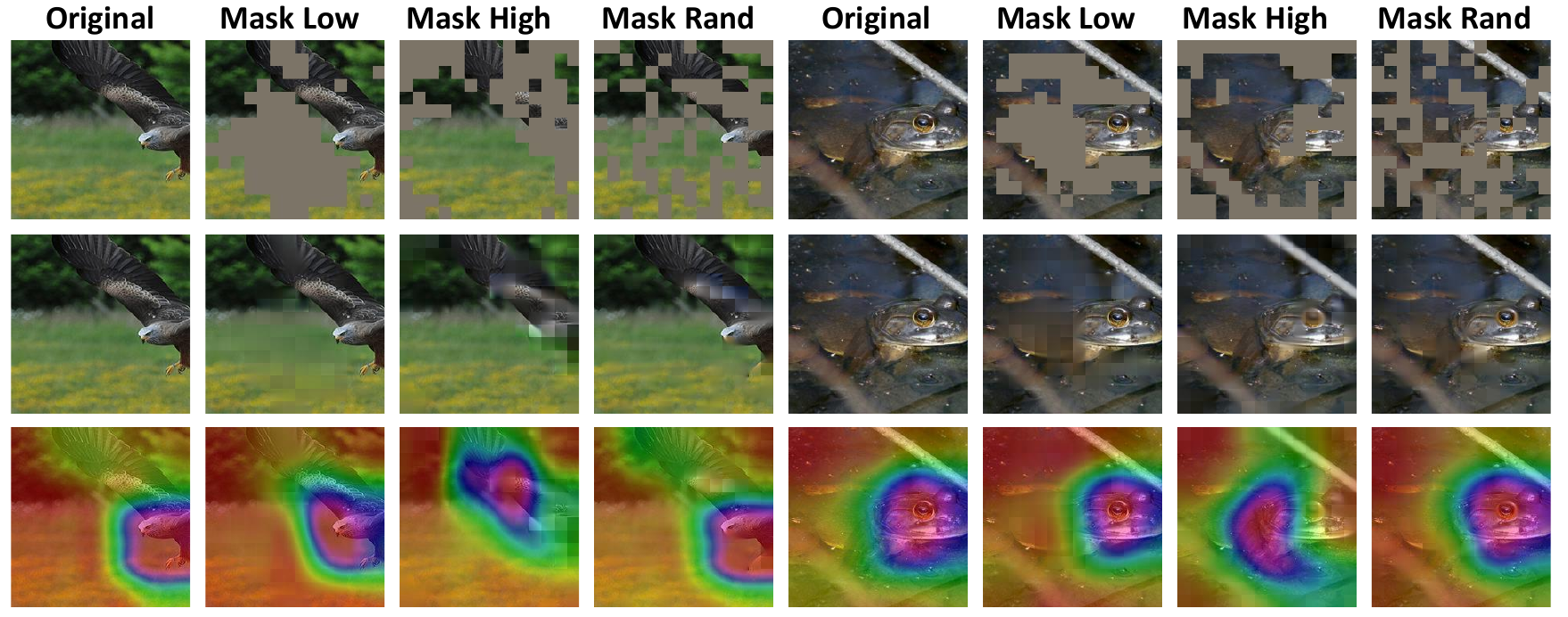}
    \caption{Visualization of different masking strategies. The first row visualizes the masked images under different masking strategy. The second row shows the reconstructed images by pretrained MAE-Mini. The gradient-weighted class activation mapping~(Grad-CAM) is shown in the last row.}
    \label{fig:mask_stra}
\end{figure}

\begin{figure*}[htp]
\centering
\begin{minipage}{.49\textwidth}
    \centering
    \includegraphics[width=.82\linewidth]{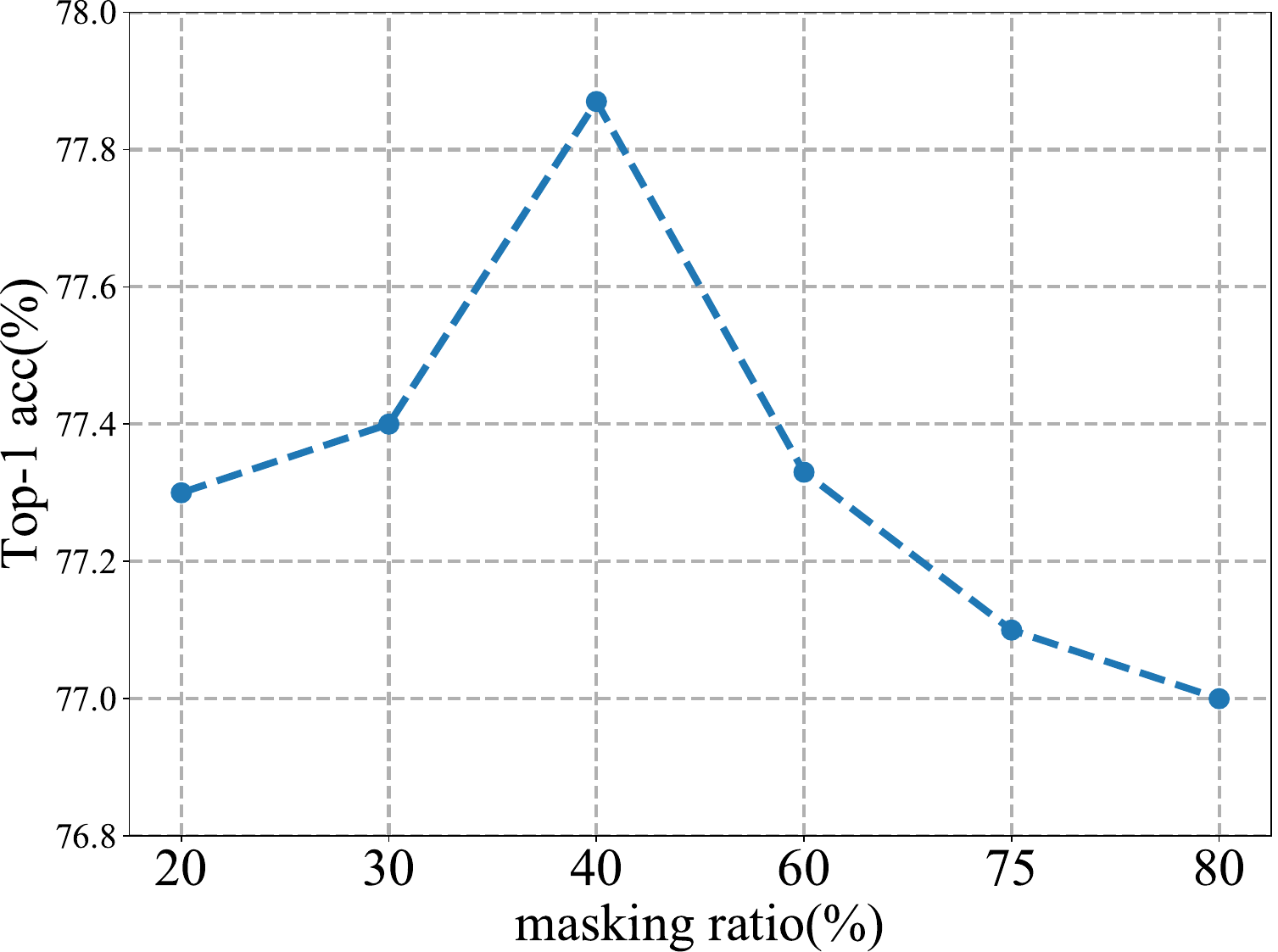}
    \vspace{-0.1in}
    \caption{Top-1 accuracy on ImageNet with different masking ratio when pretraining MAE-Mini.}
    \label{fig:maskratio}
\end{minipage}
\hspace{0.01in}
\begin{minipage}{.49\textwidth}
    \centering
    \includegraphics[width=.82\linewidth]{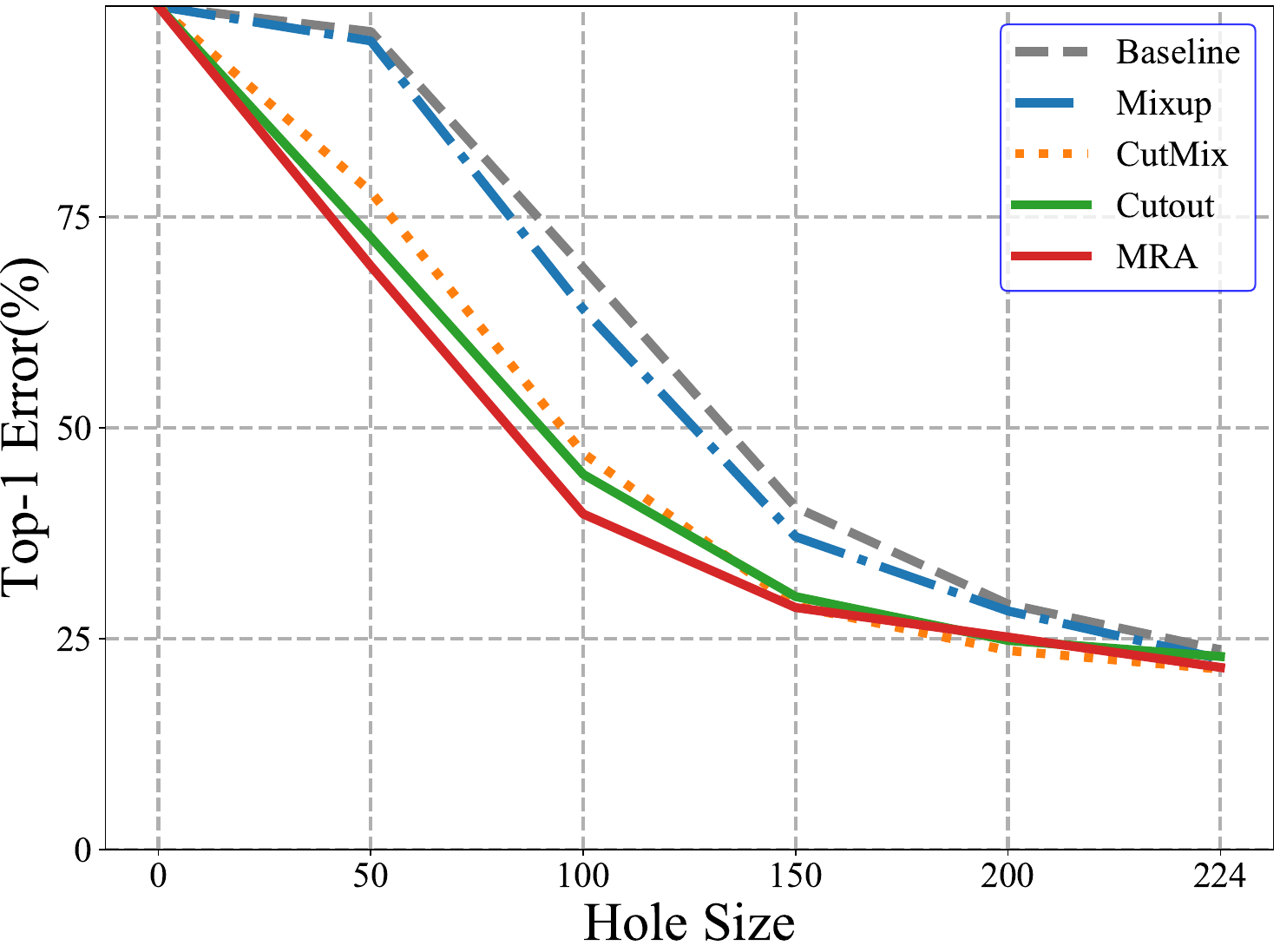}
    \vspace{-0.1in}
    \caption{Top-1 error rate on ImageNet validation set with boundary occlusion.}
    \label{fig:maskbound}
\end{minipage}
\end{figure*}

\paragraph{Masking Ratio.} To inspect how the mask ratio contributes to augmentation quality, we ablate the mask ratio ranging from 20\% to 80\%. We report the result in Figure~\ref{fig:maskratio}. It shows that the MAE-Mini pretrained under a ratio of $40\%$ reaches the best performance. We conjecture that the smaller model may not converge well with a high masking ratio. However, an extremely small masking ratio will also make the pretraining task too easy, which may influence the generalization ability of the pretrained MAE-Mini.


\paragraph{Masking Strategy.} To verify that emphasizing the semantic-related patches advances the model performance, we compare ours with other strategies where we choose to mask the patches of high attention or random patches. We report the corresponding classification accuracy in Table~\ref{tab:mask_stra}. 
It is proved that the selection of the masked area affects the performance significantly. When masking a high-attention area, the model degrades the classification performance by over 1\% compared to baseline.
Moreover, though both random masking and low-attention masking raise the accuracy, low-attention dropping rules is superior with a further nearly 0.7\% gain. Besides, as shown in Figure~\ref{fig:mask_stra}, if we erase patches of high attention like the head of the bird in the image, the reconstructed image is hard to recognize due to the vagueness of the class-specific area. It verifies that keeping patches with high attention as generation cues can produce a more robust vicinity of original image. 



\paragraph{Model Size.} We ablate the model size of MAE. As shown in Table~\ref{tab:modelsize}, under the same masking ratio, adopting a larger model as the augmentor brings higher classification accuracy. It is not surprising since the larger model captures more accurate attention information and provides stronger regularization. However, the memory and speed cost of a large MAE model is not affordable. 
By tuning the masking ratio, we show that a much smaller MAE-Mini can achieve better performance with 6$\times$ speed up and 95\% parameter decrease compared to the MAE-Large.

\begin{table}[t]
\centering
\caption{ImageNet classification accuracy under different masking strategy.}
\vspace{-0.1in}
\begin{tabular}{lcccc}
\toprule
Method   & Baseline  & Mask Low & Mask High  & Mask Random \\
\midrule
Top-1  &76.31 &77.87 & 75.24 & 77.20   \\
\bottomrule
\end{tabular}
\label{tab:mask_stra}
\vspace{-0.05in}
\end{table}


\begin{wraptable}{r}{5.2cm}
\caption{ImageNet classification accuracy with different pretraining epochs on MAE-Mini.}
\vspace{-0.005in}
\begin{tabular}{cccc}
\toprule
Epochs & 100 &  200    & 800\\
\midrule
Top-1  &77.35  &77.87 &77.76 \\
\bottomrule
\end{tabular}
\label{tab:pre-epoch}
\end{wraptable}

\begin{table}[!htp]
\centering
\caption{ImageNet classification accuracy with/without reconstruction.}
\vspace{-0.1in}
\begin{tabular}{lcccc}
\toprule
Method & Baseline &  Cutout    &\name\ Mask  &\name   \\
\midrule
Top-1  &76.31 &77.07 &77.46 &77.87 \\
\bottomrule
\end{tabular}
\label{tab:rec}
\vspace{-0.05in}
\end{table}

\begin{table}[!htp]
\small
\centering
\renewcommand\arraystretch{1.}
\setlength{\tabcolsep}{1.1mm}

\caption{ImageNet classification accuracy when using different MAE models as augmentors\protect\footnotemark[4].}
\vspace{-0.1in}
\begin{tabular}{lccccccc}
\toprule
Model      &Encoder   &Decoder   &Embed-Dim  &Mask-Ratio &Params(M) &Imgs/s  &Top-1  \\
\midrule
MAE-Base   &12        &8         &768  &0.75       &111.7  &334  &77.50  \\
MAE-Large  &12        &8         &1024 &0.75       &329.2  &160  &77.63  \\
MAE-Mini   &4         &2         &480  &0.75       &15.5   &963  &77.08  \\
MAE-Mini   &4         &2         &480  &0.40       &15.5   &963  &77.87  \\
\bottomrule
\end{tabular}
\vspace{-0.2in}
\label{tab:modelsize}
\end{table}
\footnotetext[4]{The pretrained MAE-Base and MAE-Large models are directly downloaded from https://github.com/facebookresearch/mae}

\paragraph{Pretraining Epoch.} Pretraining Epochs is an important hyper-parameter for self-supervised learning. For example, MoCo-v2 need 800 epochs, and MAE need 1600 epochs to converge with a large model. We compare the classification accuracy with \name\ under different pretraining epochs in Table \ref{tab:pre-epoch}. There is no obvious difference when extending the pretraining epochs from 200 to 800, which shows that 200 epochs pretraining is sufficient for light-weight MAE-Mini. 



\paragraph{Reconstruction.} Attention-based masking and reconstruction are two main steps in \name. To manifest the importance of the reconstruction, we design an experiment that only masks the input image. As shown in Table \ref{tab:rec}, the attention-based masking in \name\ outperforms vanilla Cutout augmentation. It is consistent with our intuition that the attention-based masking can be seen as an advanced Cutout. In addition, the performance is further improved with reconstruction, showing the effectiveness of the generation-based augmentation.

\subsection{Robustness to Occlusion}
Following CutMix~\cite{yun2019cutmix}, we further evaluate the robustness of \name\ by generating boundary occluded validation samples. Specially, we fill zeros outside of a center hole. As shown in Figure~\ref{fig:maskbound}, hole size of 0 means that the whole image is occluded, while hole size of 224 denotes that the input image is not occluded. \name\ achieves the lowest error among three augmentations which demonstrates that our mask-and-reconstruct pipeline generates occlusion-robust augmentation. 

\section{Conclusions and Limitations}
This paper proposed a robust data augmentation method, Mask-Reconstruct Augmentation (MRA) to regulate the training of deep neural networks. Via Mask-Reconstruct Augmentation, we augment the original image by reconstruction from its parts of regions. When only generating the masked regions, the augmentation can be controllable but strong because of the non-linearity. 
The experiments among a bunch of classification benchmarks demonstrate the effectiveness and robustness of MRA. 

\paragraph{Limitations:} Though our work shows promising results, there are still some limitations. This kind of augmentation is unsuitable for dense prediction tasks such as instance segmentation since generative augmentation can easily destroy the boundary of the instance. 

\paragraph{Border Impact:} This work does not have a direct negative social impact. However the powerful neural network may be used for harmful applications like face recognition.



{
\small
\bibliographystyle{plain}
\bibliography{ref}

\begin{thebibliography}{10}

\bibitem{antoniou2017data}
Antreas Antoniou, Amos Storkey, and Harrison Edwards.
\newblock Data augmentation generative adversarial networks.
\newblock {\em arXiv preprint arXiv:1711.04340}, 2017.

\bibitem{assran2022masked}
Mahmoud Assran, Mathilde Caron, Ishan Misra, Piotr Bojanowski, Florian Bordes,
  Pascal Vincent, Armand Joulin, Michael Rabbat, and Nicolas Ballas.
\newblock Masked siamese networks for label-efficient learning.
\newblock {\em arXiv preprint arXiv:2204.07141}, 2022.

\bibitem{bao2021beit}
Hangbo Bao, Li~Dong, and Furu Wei.
\newblock Beit: Bert pre-training of image transformers.
\newblock {\em arXiv preprint arXiv:2106.08254}, 2021.

\bibitem{bertalmio2000image}
Marcelo Bertalmio, Guillermo Sapiro, Vincent Caselles, and Coloma Ballester.
\newblock Image inpainting.
\newblock In {\em Proceedings of the 27th annual conference on Computer
  graphics and interactive techniques}, pages 417--424, 2000.

\bibitem{berthelot2019mixmatch}
David Berthelot, Nicholas Carlini, Ian Goodfellow, Nicolas Papernot, Avital
  Oliver, and Colin~A Raffel.
\newblock Mixmatch: A holistic approach to semi-supervised learning.
\newblock {\em Advances in Neural Information Processing Systems}, 32, 2019.

\bibitem{cai2018proxylessnas}
Han Cai, Ligeng Zhu, and Song Han.
\newblock Proxylessnas: Direct neural architecture search on target task and
  hardware.
\newblock {\em arXiv preprint arXiv:1812.00332}, 2018.

\bibitem{caron2021emerging}
Mathilde Caron, Hugo Touvron, Ishan Misra, Herv{\'e} J{\'e}gou, Julien Mairal,
  Piotr Bojanowski, and Armand Joulin.
\newblock Emerging properties in self-supervised vision transformers.
\newblock In {\em Proceedings of the IEEE/CVF International Conference on
  Computer Vision}, pages 9650--9660, 2021.

\bibitem{chen2017deeplab}
Liang-Chieh Chen, George Papandreou, Iasonas Kokkinos, Kevin Murphy, and Alan~L
  Yuille.
\newblock Deeplab: Semantic image segmentation with deep convolutional nets,
  atrous convolution, and fully connected crfs.
\newblock {\em IEEE transactions on pattern analysis and machine intelligence},
  40(4):834--848, 2017.

\bibitem{chen2018closer}
Wei-Yu Chen, Yen-Cheng Liu, Zsolt Kira, Yu-Chiang~Frank Wang, and Jia-Bin
  Huang.
\newblock A closer look at few-shot classification.
\newblock In {\em International Conference on Learning Representations}, 2018.

\bibitem{chen2022context}
Xiaokang Chen, Mingyu Ding, Xiaodi Wang, Ying Xin, Shentong Mo, Yunhao Wang,
  Shumin Han, Ping Luo, Gang Zeng, and Jingdong Wang.
\newblock Context autoencoder for self-supervised representation learning.
\newblock {\em arXiv preprint arXiv:2202.03026}, 2022.

\bibitem{ciregan2012multi}
Dan Ciregan, Ueli Meier, and J{\"u}rgen Schmidhuber.
\newblock Multi-column deep neural networks for image classification.
\newblock In {\em 2012 IEEE conference on computer vision and pattern
  recognition}, pages 3642--3649. IEEE, 2012.

\bibitem{coates2011analysis}
Adam Coates, Andrew Ng, and Honglak Lee.
\newblock An analysis of single-layer networks in unsupervised feature
  learning.
\newblock In {\em Proceedings of the fourteenth international conference on
  artificial intelligence and statistics}, pages 215--223. JMLR Workshop and
  Conference Proceedings, 2011.

\bibitem{cubuk2019autoaugment}
Ekin~D Cubuk, Barret Zoph, Dandelion Mane, Vijay Vasudevan, and Quoc~V Le.
\newblock Autoaugment: Learning augmentation strategies from data.
\newblock In {\em Proceedings of the IEEE/CVF Conference on Computer Vision and
  Pattern Recognition}, pages 113--123, 2019.

\bibitem{cubuk2020randaugment}
Ekin~D Cubuk, Barret Zoph, Jonathon Shlens, and Quoc~V Le.
\newblock Randaugment: Practical automated data augmentation with a reduced
  search space.
\newblock In {\em Proceedings of the IEEE/CVF Conference on Computer Vision and
  Pattern Recognition Workshops}, pages 702--703, 2020.

\bibitem{deng2009imagenet}
Jia Deng, Wei Dong, Richard Socher, Li-Jia Li, Kai Li, and Li~Fei-Fei.
\newblock Imagenet: A large-scale hierarchical image database.
\newblock In {\em 2009 IEEE conference on computer vision and pattern
  recognition}, pages 248--255. Ieee, 2009.

\bibitem{devries2017improved}
Terrance DeVries and Graham~W Taylor.
\newblock Improved regularization of convolutional neural networks with cutout.
\newblock {\em arXiv preprint arXiv:1708.04552}, 2017.

\bibitem{dosovitskiy2020image}
Alexey Dosovitskiy, Lucas Beyer, Alexander Kolesnikov, Dirk Weissenborn,
  Xiaohua Zhai, Thomas Unterthiner, Mostafa Dehghani, Matthias Minderer, Georg
  Heigold, Sylvain Gelly, et~al.
\newblock An image is worth 16x16 words: Transformers for image recognition at
  scale.
\newblock In {\em International Conference on Learning Representations}, 2020.

\bibitem{fang2022corrupted}
Yuxin Fang, Li~Dong, Hangbo Bao, Xinggang Wang, and Furu Wei.
\newblock Corrupted image modeling for self-supervised visual pre-training.
\newblock {\em arXiv preprint arXiv:2202.03382}, 2022.

\bibitem{finn2017model}
Chelsea Finn, Pieter Abbeel, and Sergey Levine.
\newblock Model-agnostic meta-learning for fast adaptation of deep networks.
\newblock In {\em International conference on machine learning}, pages
  1126--1135. PMLR, 2017.

\bibitem{frid2018gan}
Maayan Frid-Adar, Idit Diamant, Eyal Klang, Michal Amitai, Jacob Goldberger,
  and Hayit Greenspan.
\newblock Gan-based synthetic medical image augmentation for increased cnn
  performance in liver lesion classification.
\newblock {\em Neurocomputing}, 321:321--331, 2018.

\bibitem{gidaris2018unsupervised}
Spyros Gidaris, Praveer Singh, and Nikos Komodakis.
\newblock Unsupervised representation learning by predicting image rotations.
\newblock In {\em International Conference on Learning Representations}, 2018.

\bibitem{goodfellow2014generative}
Ian Goodfellow, Jean Pouget-Abadie, Mehdi Mirza, Bing Xu, David Warde-Farley,
  Sherjil Ozair, Aaron Courville, and Yoshua Bengio.
\newblock Generative adversarial nets.
\newblock {\em Advances in neural information processing systems}, 27, 2014.

\bibitem{hataya2020faster}
Ryuichiro Hataya, Jan Zdenek, Kazuki Yoshizoe, and Hideki Nakayama.
\newblock Faster autoaugment: Learning augmentation strategies using
  backpropagation.
\newblock In {\em European Conference on Computer Vision}, pages 1--16.
  Springer, 2020.

\bibitem{he2021masked}
Kaiming He, Xinlei Chen, Saining Xie, Yanghao Li, Piotr Doll{\'a}r, and Ross
  Girshick.
\newblock Masked autoencoders are scalable vision learners.
\newblock {\em arXiv preprint arXiv:2111.06377}, 2021.

\bibitem{he2016deep}
Kaiming He, Xiangyu Zhang, Shaoqing Ren, and Jian Sun.
\newblock Deep residual learning for image recognition.
\newblock In {\em Proceedings of the IEEE conference on computer vision and
  pattern recognition}, pages 770--778, 2016.

\bibitem{isola2017image}
Phillip Isola, Jun-Yan Zhu, Tinghui Zhou, and Alexei~A Efros.
\newblock Image-to-image translation with conditional adversarial networks.
\newblock In {\em Proceedings of the IEEE conference on computer vision and
  pattern recognition}, pages 1125--1134, 2017.

\bibitem{kang2019decoupling}
Bingyi Kang, Saining Xie, Marcus Rohrbach, Zhicheng Yan, Albert Gordo, Jiashi
  Feng, and Yannis Kalantidis.
\newblock Decoupling representation and classifier for long-tailed recognition.
\newblock In {\em International Conference on Learning Representations}, 2019.

\bibitem{kim2020puzzle}
Jang-Hyun Kim, Wonho Choo, and Hyun~Oh Song.
\newblock Puzzle mix: Exploiting saliency and local statistics for optimal
  mixup.
\newblock In {\em International Conference on Machine Learning}, pages
  5275--5285. PMLR, 2020.

\bibitem{kingma2013auto}
Diederik~P Kingma and Max Welling.
\newblock Auto-encoding variational bayes.
\newblock {\em arXiv preprint arXiv:1312.6114}, 2013.

\bibitem{krause20133d}
Jonathan Krause, Michael Stark, Jia Deng, and Li~Fei-Fei.
\newblock 3d object representations for fine-grained categorization.
\newblock In {\em Proceedings of the IEEE international conference on computer
  vision workshops}, pages 554--561, 2013.

\bibitem{krizhevsky2012imagenet}
Alex Krizhevsky, Ilya Sutskever, and Geoffrey~E Hinton.
\newblock Imagenet classification with deep convolutional neural networks.
\newblock {\em Advances in neural information processing systems}, 25, 2012.

\bibitem{lecun1998gradient}
Yann LeCun, L{\'e}on Bottou, Yoshua Bengio, and Patrick Haffner.
\newblock Gradient-based learning applied to document recognition.
\newblock {\em Proceedings of the IEEE}, 86(11):2278--2324, 1998.

\bibitem{lee2013pseudo}
Dong-Hyun Lee et~al.
\newblock Pseudo-label: The simple and efficient semi-supervised learning
  method for deep neural networks.
\newblock In {\em Workshop on challenges in representation learning, ICML},
  volume~3, page 896, 2013.

\bibitem{li2020dada}
Yonggang Li, Guosheng Hu, Yongtao Wang, Timothy Hospedales, Neil~M Robertson,
  and Yongxin Yang.
\newblock Dada: differentiable automatic data augmentation.
\newblock {\em arXiv preprint arXiv:2003.03780}, 2020.

\bibitem{lim2019fast}
Sungbin Lim, Ildoo Kim, Taesup Kim, Chiheon Kim, and Sungwoong Kim.
\newblock Fast autoaugment.
\newblock {\em Advances in Neural Information Processing Systems}, 32, 2019.

\bibitem{liu2022convnet}
Zhuang Liu, Hanzi Mao, Chao-Yuan Wu, Christoph Feichtenhofer, Trevor Darrell,
  and Saining Xie.
\newblock A convnet for the 2020s.
\newblock {\em arXiv preprint arXiv:2201.03545}, 2022.

\bibitem{long2015fully}
Jonathan Long, Evan Shelhamer, and Trevor Darrell.
\newblock Fully convolutional networks for semantic segmentation.
\newblock In {\em Proceedings of the IEEE conference on computer vision and
  pattern recognition}, pages 3431--3440, 2015.

\bibitem{madani2018chest}
Ali Madani, Mehdi Moradi, Alexandros Karargyris, and Tanveer Syeda-Mahmood.
\newblock Chest x-ray generation and data augmentation for cardiovascular
  abnormality classification.
\newblock In {\em Medical Imaging 2018: Image Processing}, volume 10574, page
  105741M. International Society for Optics and Photonics, 2018.

\bibitem{maji2013fine}
Subhransu Maji, Esa Rahtu, Juho Kannala, Matthew Blaschko, and Andrea Vedaldi.
\newblock Fine-grained visual classification of aircraft.
\newblock {\em arXiv preprint arXiv:1306.5151}, 2013.

\bibitem{mikolajczyk2018data}
Agnieszka Miko{\l}ajczyk and Micha{\l} Grochowski.
\newblock Data augmentation for improving deep learning in image classification
  problem.
\newblock In {\em 2018 international interdisciplinary PhD workshop (IIPhDW)},
  pages 117--122. IEEE, 2018.

\bibitem{miyato2018virtual}
Takeru Miyato, Shin-ichi Maeda, Masanori Koyama, and Shin Ishii.
\newblock Virtual adversarial training: a regularization method for supervised
  and semi-supervised learning.
\newblock {\em IEEE transactions on pattern analysis and machine intelligence},
  41(8):1979--1993, 2018.

\bibitem{pathak2016context}
Deepak Pathak, Philipp Krahenbuhl, Jeff Donahue, Trevor Darrell, and Alexei~A
  Efros.
\newblock Context encoders: Feature learning by inpainting.
\newblock In {\em Proceedings of the IEEE conference on computer vision and
  pattern recognition}, pages 2536--2544, 2016.

\bibitem{radford2015unsupervised}
Alec Radford, Luke Metz, and Soumith Chintala.
\newblock Unsupervised representation learning with deep convolutional
  generative adversarial networks.
\newblock {\em arXiv preprint arXiv:1511.06434}, 2015.

\bibitem{rasmus2015semi}
Antti Rasmus, Mathias Berglund, Mikko Honkala, Harri Valpola, and Tapani Raiko.
\newblock Semi-supervised learning with ladder networks.
\newblock {\em Advances in neural information processing systems}, 28, 2015.

\bibitem{redmon2016you}
Joseph Redmon, Santosh Divvala, Ross Girshick, and Ali Farhadi.
\newblock You only look once: Unified, real-time object detection.
\newblock In {\em Proceedings of the IEEE conference on computer vision and
  pattern recognition}, pages 779--788, 2016.

\bibitem{ren2015faster}
Shaoqing Ren, Kaiming He, Ross Girshick, and Jian Sun.
\newblock Faster r-cnn: Towards real-time object detection with region proposal
  networks.
\newblock {\em Advances in neural information processing systems}, 28, 2015.

\bibitem{schwartz2018delta}
Eli Schwartz, Leonid Karlinsky, Joseph Shtok, Sivan Harary, Mattias Marder,
  Abhishek Kumar, Rogerio Feris, Raja Giryes, and Alex Bronstein.
\newblock Delta-encoder: an effective sample synthesis method for few-shot
  object recognition.
\newblock {\em Advances in Neural Information Processing Systems}, 31, 2018.

\bibitem{shorten2019survey}
Connor Shorten and Taghi~M Khoshgoftaar.
\newblock A survey on image data augmentation for deep learning.
\newblock {\em Journal of big data}, 6(1):1--48, 2019.

\bibitem{simonyan2014very}
Karen Simonyan and Andrew Zisserman.
\newblock Very deep convolutional networks for large-scale image recognition.
\newblock {\em arXiv preprint arXiv:1409.1556}, 2014.

\bibitem{snell2017prototypical}
Jake Snell, Kevin Swersky, and Richard Zemel.
\newblock Prototypical networks for few-shot learning.
\newblock {\em Advances in neural information processing systems}, 30, 2017.

\bibitem{sohn2020fixmatch}
Kihyuk Sohn, David Berthelot, Nicholas Carlini, Zizhao Zhang, Han Zhang,
  Colin~A Raffel, Ekin~Dogus Cubuk, Alexey Kurakin, and Chun-Liang Li.
\newblock Fixmatch: Simplifying semi-supervised learning with consistency and
  confidence.
\newblock {\em Advances in Neural Information Processing Systems}, 33:596--608,
  2020.

\bibitem{sung2018learning}
Flood Sung, Yongxin Yang, Li~Zhang, Tao Xiang, Philip~HS Torr, and Timothy~M
  Hospedales.
\newblock Learning to compare: Relation network for few-shot learning.
\newblock In {\em Proceedings of the IEEE conference on computer vision and
  pattern recognition}, pages 1199--1208, 2018.

\bibitem{tarvainen2017mean}
Antti Tarvainen and Harri Valpola.
\newblock Mean teachers are better role models: Weight-averaged consistency
  targets improve semi-supervised deep learning results.
\newblock {\em Advances in neural information processing systems}, 30, 2017.

\bibitem{touvron2021training}
Hugo Touvron, Matthieu Cord, Matthijs Douze, Francisco Massa, Alexandre
  Sablayrolles, and Herv{\'e} J{\'e}gou.
\newblock Training data-efficient image transformers \& distillation through
  attention.
\newblock In {\em International Conference on Machine Learning}, pages
  10347--10357. PMLR, 2021.

\bibitem{uddin2020saliencymix}
AFM~Shahab Uddin, Mst~Sirazam Monira, Wheemyung Shin, TaeChoong Chung, and
  Sung-Ho Bae.
\newblock Saliencymix: A saliency guided data augmentation strategy for better
  regularization.
\newblock In {\em International Conference on Learning Representations}, 2020.

\bibitem{vinyals2016matching}
Oriol Vinyals, Charles Blundell, Timothy Lillicrap, Daan Wierstra, et~al.
\newblock Matching networks for one shot learning.
\newblock {\em Advances in neural information processing systems}, 29, 2016.

\bibitem{wah2011caltech}
Catherine Wah, Steve Branson, Peter Welinder, Pietro Perona, and Serge
  Belongie.
\newblock The caltech-ucsd birds-200-2011 dataset.
\newblock 2011.

\bibitem{wan2013regularization}
Li~Wan, Matthew Zeiler, Sixin Zhang, Yann Le~Cun, and Rob Fergus.
\newblock Regularization of neural networks using dropconnect.
\newblock In {\em International conference on machine learning}, pages
  1058--1066. PMLR, 2013.

\bibitem{xie2020unsupervised}
Qizhe Xie, Zihang Dai, Eduard Hovy, Thang Luong, and Quoc Le.
\newblock Unsupervised data augmentation for consistency training.
\newblock {\em Advances in Neural Information Processing Systems},
  33:6256--6268, 2020.

\bibitem{xie2017aggregated}
Saining Xie, Ross Girshick, Piotr Doll{\'a}r, Zhuowen Tu, and Kaiming He.
\newblock Aggregated residual transformations for deep neural networks.
\newblock In {\em Proceedings of the IEEE conference on computer vision and
  pattern recognition}, pages 1492--1500, 2017.

\bibitem{yi2019generative}
Xin Yi, Ekta Walia, and Paul Babyn.
\newblock Generative adversarial network in medical imaging: A review.
\newblock {\em Medical image analysis}, 58:101552, 2019.

\bibitem{yun2019cutmix}
Sangdoo Yun, Dongyoon Han, Seong~Joon Oh, Sanghyuk Chun, Junsuk Choe, and
  Youngjoon Yoo.
\newblock Cutmix: Regularization strategy to train strong classifiers with
  localizable features.
\newblock In {\em Proceedings of the IEEE/CVF international conference on
  computer vision}, pages 6023--6032, 2019.

\bibitem{zagoruyko2016wide}
Sergey Zagoruyko and Nikos Komodakis.
\newblock Wide residual networks.
\newblock {\em arXiv preprint arXiv:1605.07146}, 2016.

\bibitem{zhang2021understanding}
Chiyuan Zhang, Samy Bengio, Moritz Hardt, Benjamin Recht, and Oriol Vinyals.
\newblock Understanding deep learning (still) requires rethinking
  generalization.
\newblock {\em Communications of the ACM}, 64(3):107--115, 2021.

\bibitem{zhang2018mixup}
Hongyi Zhang, Moustapha Cisse, Yann~N Dauphin, and David Lopez-Paz.
\newblock mixup: Beyond empirical risk minimization.
\newblock In {\em International Conference on Learning Representations}, 2018.

\bibitem{zhang2019aet}
Liheng Zhang, Guo-Jun Qi, Liqiang Wang, and Jiebo Luo.
\newblock Aet vs. aed: Unsupervised representation learning by auto-encoding
  transformations rather than data.
\newblock In {\em Proceedings of the IEEE/CVF Conference on Computer Vision and
  Pattern Recognition}, pages 2547--2555, 2019.

\bibitem{zhong2020random}
Zhun Zhong, Liang Zheng, Guoliang Kang, Shaozi Li, and Yi~Yang.
\newblock Random erasing data augmentation.
\newblock In {\em Proceedings of the AAAI conference on artificial
  intelligence}, volume~34, pages 13001--13008, 2020.

\bibitem{zhu2018emotion}
Xinyue Zhu, Yifan Liu, Jiahong Li, Tao Wan, and Zengchang Qin.
\newblock Emotion classification with data augmentation using generative
  adversarial networks.
\newblock In {\em Pacific-Asia conference on knowledge discovery and data
  mining}, pages 349--360. Springer, 2018.

\end{thebibliography}
}

\end{document}